\pdfoutput=1

\documentclass[11pt]{article}

\usepackage[final]{acl}

\usepackage{times}
\usepackage{latexsym}

\usepackage[T1]{fontenc}

\usepackage{helvet}  
\usepackage{courier}  

\usepackage{natbib}  
\usepackage{caption} 
\usepackage{times}  
\usepackage{helvet}  
\usepackage{courier}  
\usepackage{natbib}  
\usepackage{caption} 
\usepackage{amsmath}
\usepackage{algpseudocode}
\usepackage{algorithm}
\usepackage{array}
\usepackage{colortbl}
\usepackage{hhline}
\usepackage[T1]{fontenc}
\usepackage{rotating}
\usepackage{microtype}
\usepackage{xcolor}
\usepackage{multirow}
\usepackage[utf8]{inputenc} 
\usepackage{url}            
\usepackage{booktabs}       
\usepackage{amsfonts}       
\usepackage{nicefrac}       
\usepackage{times}
\usepackage{latexsym}
\usepackage{booktabs} 
\usepackage{tabularx}
\usepackage{subcaption} 
\usepackage[T1]{fontenc}
\usepackage{microtype}
\usepackage{inconsolata}

\usepackage{bibentry}
\usepackage{newfloat}
\usepackage{listings}
\DeclareCaptionStyle{ruled}{labelfont=normalfont,labelsep=colon,strut=off} 
\floatstyle{ruled}
\newfloat{listing}{tb}{lst}{}
\floatname{listing}{Listing}
\title{TUNI: A Textual Unimodal Detector for Identity Inference in CLIP Models}

\author{
  Songze Li$^{1, *, \dagger}$, 
  Ruoxi Cheng$^{1,}$\thanks{
  Contributed equally to this work. 
  $^1$Southeast University, Nanjing China.
  $^2$Nanyang Technological University, Singapore.
  $^{\dagger}$Corresponding authors: \href{mailto:songzeli@seu.edu.cn}{songzeli@seu.edu.cn}.
  },
  Xiaojun Jia$^{2}$
}

\begin{document}

\maketitle

\begin{abstract}

The widespread usage of large-scale multimodal models like CLIP has heightened concerns about the leakage of PII. Existing methods for identity inference in CLIP models require querying the model with full PII, including textual descriptions of the person and corresponding images (e.g., the name and the face photo of the person). However, applying images may risk exposing personal information to target models, as the image might not have been previously encountered by the target model.
Additionally, previous MIAs train shadow models to mimic the behaviors of the target model, which incurs high computational costs, especially for large CLIP models. To address these challenges, we propose a textual unimodal detector (TUNI) in CLIP models, a novel technique for identity inference that: 1) only utilizes text data to query the target model; and 2) eliminates the need for training shadow models. Extensive experiments of TUNI across various CLIP model architectures and datasets demonstrate its superior performance over baselines, albeit with only text data. 

\end{abstract}

\section{Introduction}

Recent years have witnessed a rapid development of large-scale multimodal models, such as Contrastive Language–Image Pre-training (CLIP)~\citep{radford2021learning}. These models synthesize information across different modalities, particularly text and images, facilitating applications from automated image generation to sophisticated visual question answering systems. Despite their potential, these models pose significant privacy risks~\citep{inan2021training,carlini2021extracting,leino2020stolen,rigaki2023survey,helbling2023llm,rahman2024survey,rahman2023survey} as the vast datasets used for training often contain personally identifiable information (PII)~\citep{schwartz2011pii,abadi2016deep,bonawitz2017practical}, raising concerns~\citep{xi2024defending} about PII leakage and misuse~\citep{hu2023defenses,yin2021defending}. Therefore, it is extremely important to develop tools to detect potential PII leakage from CLIP models. Specially, as the first step, we would like to address the identity inference problem, i.e., to determine if the PII of a particular person was used in training of a target CLIP model.

\begin{figure*}[t]
  \centering
  \includegraphics[width=\textwidth]{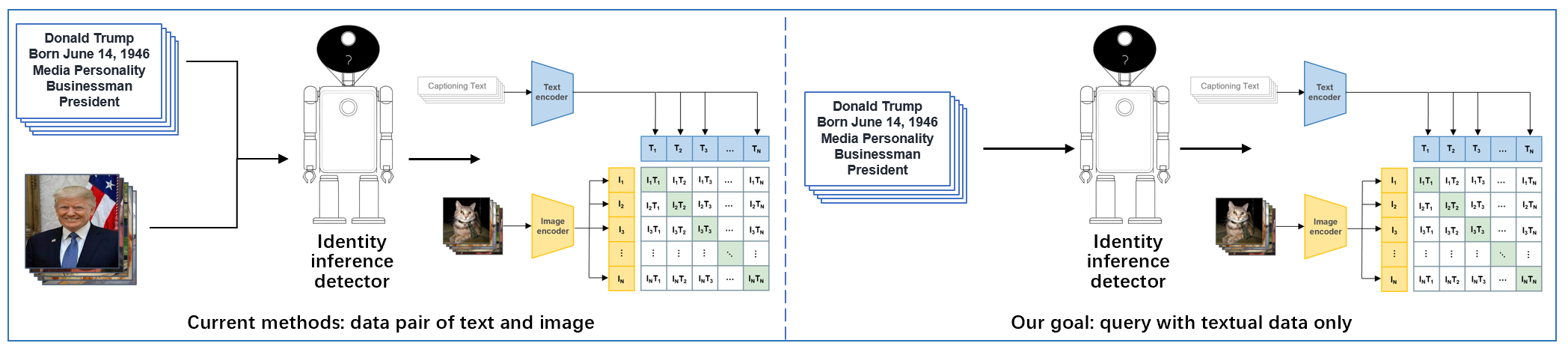}
  \caption{\small Current methods query LLMs with both text and image, while our goal is to conduct identity inference with only textual data.}
  \label{fig:example}
\end{figure*}

Traditional methods, like Membership Inference Attacks (MIAs)~\citep{shokri2017membership}, have focused on determining whether a specific data sample was used for model training. When applied to CLIP models, these approaches typically involve querying the model with
both texts and images of the target individual~\citep{ko2023}, and exposing images of a person the CLIP model may have not seen in the training set brings new privacy leakage risk~\citep{he2022membership}. Hence, it is desirable to have a detection mechanism for ID inference that \emph{does not query the CLIP model with real images of the person} (see an example in Figure~\ref{fig:example}). Furthermore, traditional MIAs often rely on constructing shadow models that mimic the behaviors of the target model to obtain training data to construct attack models ~\citep{hu2022membership}, which demands extensive computational resources and is less feasible in environments with limited computational capabilities~\citep{mattern2023membership,hisamoto2020membership,jagielski2024students}. Alternative methods for shadow models in MIAs, such as those based on cosine similarity~\citep{ko2023} and self-influence functions~\citep{cohen2024membership}, exhibit either lower accuracy or still necessitate substantial computational resources~\citep{oh2023membership}.

 \begin{figure}[h]
    \centering
    \begin{minipage}{0.48\textwidth}
        \includegraphics[width=\linewidth]{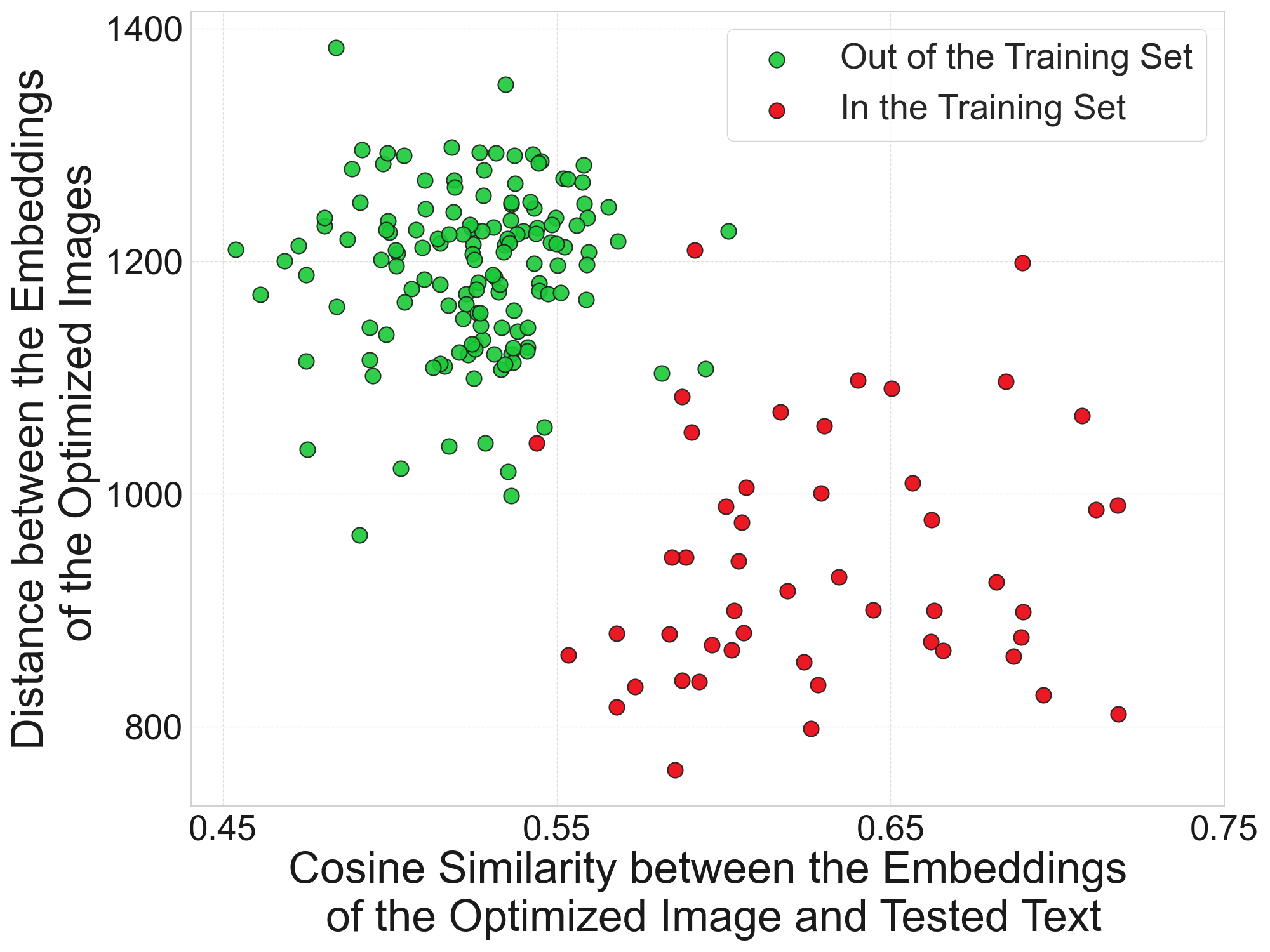}
    \end{minipage}\hfill
    \begin{minipage}{0.48\textwidth}
        \caption{\small Features of textual descriptions extracted from the optimized images guided by a CLIP model with ResNet50x4 architecture, trained on a dataset where each person has 75 images. The cosine similarity between the embeddings of optimized image and the tested text, and the distance between the embeddings of the optimized images, can clearly distinguish between the samples within and outside the training dataset of the target CLIP model.}
        \label{fig:separation}
    \end{minipage}
\end{figure}

 To address these limitations, we propose a textual unimodel detector (TUNI) for identity inference in CLIP models, which queries the target model with only text information during inference.
 Specifically, we first propose a feature extractor, which maps a textual description to a feature vector through image optimization guided by the CLIP model; then, we randomly generate a large amount of textual gibberish, which we know do not match any textual descriptions in the training dataset. 
 As shown in Figure~\ref{fig:separation}, we make the key observation that the feature distributions of textual gibberish and member samples in the training set are well distinguishable.

Leveraging this property, we use the feature vectors of the generated textual gibberish to train multiple anomaly detectors to form an anomaly detection voting system. At test time, TUNI simply feeds the feature vector of the test text to the voting system, and determines that if the corresponding PII is included in the training set (abnormal) or not (normal).
 The training of the anomaly detector in TUNI costs only several hours with four NVIDIA GeForce RTX 3090 GPUs, avoiding training shadow models with the size of the CLIP model in traditional MIAs, which can cost over 18 days even with hundreds of advanced GPUs~\citep{gu2022wukong,ko2023,hu2022m}.  


Our contributions are summarized as follows:
\vspace{-2mm}
\begin{itemize}
\item We propose a textual unimodal detector, dubbed \textit{TUNI}, which is the first method to conduct identity inference in CLIP models with unimodal data, preventing risky exposure of images to the target model;
\vspace{-2mm}
\item We find that the feature distributions of texts that are in and out of the target CLIP model are well separated, and propose to adopt randomly generated text to train anomaly detectors for ID inference, avoiding the need for computationally intensive shadow models in traditional MIAs. 
\vspace{-2mm}
\item Extensive experiments conducted across six kinds of CLIP models have indicated that the proposed TUNI achieves better performance than current methods for identity inference, even when using only textual data.

\end{itemize}


\section{Related Work}

\subsection{Privacy Leakage in CLIP Models}

CLIP model exemplifies modern multimodal innovation by integrating an image encoder and a text encoder into its architecture~\citep{radford2021learning}. These encoders transform inputs into a shared embedding space, enabling effective measurement of semantic similarity~\citep{ramesh2022hierarchical}. Despite the significant advances and expansive applicability of CLIP models, the vast and diverse datasets utilized for training such models could potentially include sensitive information, raising concerns about privacy leakage~\citep{hu2022m}. Various inference attacks, including model stealing ~\citep{dziedzic2022dataset,liu2022stolenencoder,wu2022model}, knowledge stealing~\citep{liang2022imitated}, data stealing~\citep{10.1145/3460120.3484571}, and membership inference attacks~\citep{liu2021encodermi,ko2023}, have been developed for CLIP, exposing potential vulnerability in privacy leakage. 
These privacy concerns underscore the necessity for developing robust defense mechanisms to safeguard sensitive information in CLIP models~\citep{golatkar2022mixed,jia202310,huang2023safeguarding}.

\subsection{Personally Identifiable Information and Leakage Issues}
Personally Identifiable Information (PII) is defined as any data that can either independently or when combined with other information, identify an individual. Training Large Language Models (LLMs) often utilizes publicly accessible datasets, which may inadvertently contain PII. This elevates the risk of data breaches that could compromise individual privacy and entail severe legal and reputational consequences for the deploying entities~\citep{lukas2023analyzing,abadi2016deep,bonawitz2017practical,rahman2020everything,shamshad2023clip2protect}. Various attacks have been developed to reveal PII from LLMs. A method is proposed in~\citep{panda2024teach} to steal private information from LLMs via crafting specific queries to GPT-4 that can reveal sensitive data by appending a secret suffix to the generated text; Zhang et al. introduced the ETHICIST method for targeted training data extraction, through loss smoothed soft prompting and calibrated confidence estimation, significantly improving extraction performance on public benchmarks~\citep{zhang-etal-2023-ethicist}; Carlini et al. also studied training data extraction from LLMs, emphasizing the predictive capability of attacks given a prefix~\citep{carlini2021extracting}; ProPILE, proposed in~\citep{kim2024}, probes privacy leakage in LLMs, by assessing the leakage risk of PII included in the publicly available Pile dataset; 
Inan et al. investigated the risks associated with membership inference attacks using a Reddit dataset, further emphasizing the persistent threat of PII leakage in various data environments~\citep{inan2021training}. 

\subsection{Current Identity Inference Methods and Their Limitations}
Identity inference, critical in privacy-preserving data analysis, has garnered significant attention across domains, such as genomic data~\citep{erlich2018identity}, location-based spatial queries~\citep{kalnis2007preventing}, person re-identification scenarios~\citep{karaman2012identity}, computer-mediated communication~\citep{motahari2009identity}and face recognition~\citep{zhou2018age,prince2011probabilistic,sanderson2009multi}. 
Membership Inference Attacks (MIAs), which determine if specific data points were in a model's training dataset, can be used to perform identity inference. Traditional MIAs often require constructing shadow models to mimic the target model’s behavior, posing computational efficiency challenges for large models~\citep{truex2019demystifying,ye2022enhanced,meeus2023did,9806361}. 

While identity inference has been mainly performed on unimodal models, it is recently extended to CLIP models. Identity Detection Inference Attack (IDIA)~\citep{hintersdorf2022} does not need shadow models; it involves providing real photos of the tested individual and 1000 prompt templates including the real name to choose from. The attacker generates multiple queries by substituting the <NAME> placeholder and analyzes the model’s responses to calculate an attack score based on correct predictions. If the correct name is predicted for a threshold number of templates, the individual is inferred to be in the training data. Cosine Similarity Attacks (CSA)~\citep{ko2023} uses cosine similarity (CS) between image and text features to infer membership, as CLIP is trained to maximize CS for training samples. Based on CSA, Weak Supervision Attack (WSA) uses a new weak supervision MIA framework with unilateral non-member information for enhancement. Both IDIA and WSA avoid the high costs associated with shadow models, but require querying the target model with real images the model may have never seen, raising new privacy concerns.

\section{Methodology}
\subsection{Problem Setup and Threat Model}
Consider a CLIP model \(M\) trained on a dataset \(D_{\text{train}}\). Each sample \(s_i = (t_i, x_i)\) in \(D_{\text{train}}\) records the personally identifiable information (PII) of an individual person, and consists of a textual description \(t_i\) (e.g., name of the person) and a corresponding image \(x_i\) (e.g., face photo of the person). For distinct indices \(i \neq j\), it is possible that \(t_i = t_j\) and \(x_i \neq x_j\), indicating that multiple non-identical images of the same person may exist. 

A detector would like to probe potential leakage of a person's PII through the target CLIP model $M$, via conducting an identity inference task against $M$, to determine if any PII samples of this person were included in the training set \(D_{\text{train}}\).

\textbf{Detector's Goal.} For a person with textual description \(t\), a detector would like to determine whether there exits a PII sample $(t_i,x_i) \in D_{\text{train}}$, such that $t_i=t$.

Note that rather than detecting for a particular text-image pair $(t,x)$, our goal is to detect existence of \emph{any} (one or more) pair with a textual description of $t$. This is because that 
multiple images of the same person can be used for training, and any one of these images may lead to potential PII leakage.

\textbf{Detector’s Knowledge and Capability.} The detector can query \(M\) and observe the output, including extracted image and text embeddings as well as their matching score, but does not know the model architecture of \(M\), the parameter values, or the training algorithms. For the target textual description $t$, depending on the application scenarios, the detector may or may not have actual images corresponding to $t$. \emph{Nevertheless, in the case where the detector knows corresponding images, due to privacy concerns, it cannot include them in the queries to \(M\).} The detector cannot modify \(M\) or access its internal state.

\subsection{TUNI: Textual Unimodal Detector for ID Inference}

\begin{figure*}[ht]
  \centering
  \includegraphics[width=0.9 \textwidth]{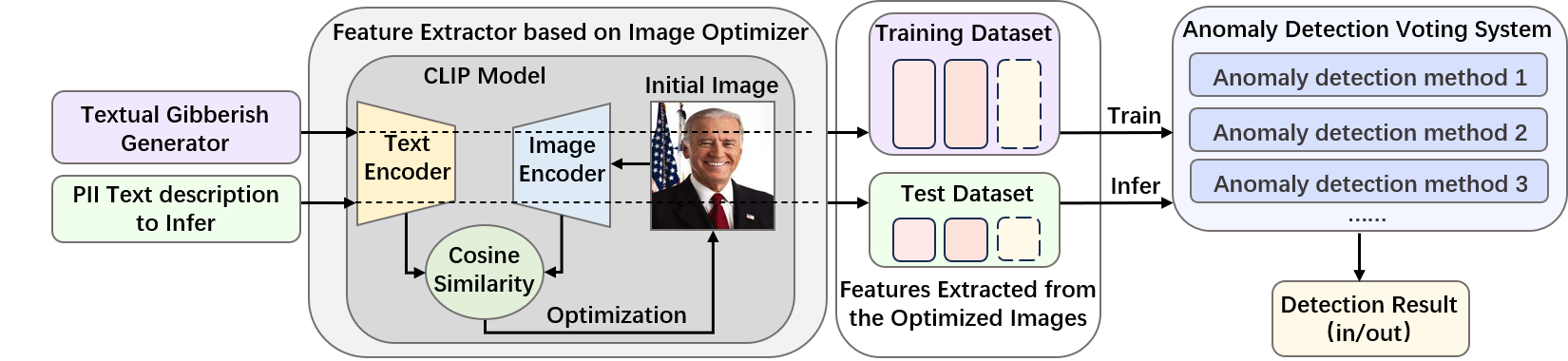}
  \caption{\small Overview of TUNI.}
  \label{fig:tuni}
\end{figure*}

We design a textual unimodal detector for ID inference (TUNI), to determine whether the PII of a person is in the training set of the target CLIP model $M$, with the restriction that only the textual description of the person can be exposed to $M$. Firstly, for a textual description $t$, we develop a feature extractor to map $t$ to a feature vector, through image optimization guided by the CLIP model. Then, we make the key observation that \emph{textual gibberish like ``D2;l-NOXRT''—random combinations of numbers and symbols clearly do not match any textual descriptions in the training set}, and hence the detector can generate large amount of textual gibberish that are known out of $D_{\text{train}}$. Using feature vectors extracted from these textual gibberish, the detector can train multiple anomaly detectors to form an anomaly detection voting system. Finally, during the inference phase, the features of the target textual description are fed into the system, and the inference result is determined through voting. Additionally, when the actual images of the textual description is available to the detector, they can be leverage to perform clustering on the feature vectors of the test samples to further enhance detection performance. An overview of the proposed TUNI framework is shown in Figure~\ref{fig:tuni}.

\textbf{Feature Extraction through CLIP-guided Image Optimization.} The feature extraction for a textual description $t$ involves iterative optimization of an image $x$, to maximize the correlation between the embeddings of $t$ and $x$ out of the target CLIP model. The extraction process, described in Algorithm~\ref{alg:feature-extraction}, 
iterates for $n$ epochs; and within each epoch, an image is optimized for $m$ iterations, to maximize the cosine similarity between its embedding of the CLIP model and that of the target textual description. The average optimized cosine similarity $S$ and standard deviation of the optimized image embeddings $D$ are extracted as the features of $t$ from model $M$.


\begin{algorithm}[tb]
\caption{CLIP-guided Feature Extraction}
\label{alg:feature-extraction}
\textbf{Input}: Target CLIP model \( M \), textual description \( t \) \\
\textbf{Output}: Mean optimized cosine similarity \( S \), standard deviation of optimized image embeddings \( D \)
\begin{algorithmic}[1]
\State $n \gets$ number of epochs
\State $m \gets$ number of optimization iterations per epoch
\State $\mathcal{S} \gets \emptyset$, $\mathcal{V} \gets \emptyset$
\State $v_t \gets M(t)$ \Comment{Obtain text embedding from $M$}
\For{$i = 1$ \textbf{to} $n$}
    \State $x_0 \gets \textup{Rand}()$ \Comment{Randomly generate an initial image}
    \For{$j = 0$ \textbf{to} $m-1$}
        \State $v_{x_j} \gets M(x_j)$ \Comment{Obtain image embedding from $M$}
        \State $x_{j+1} \gets \arg \max_{x_j} \frac{v_t \cdot v_{x_j}}{\|v_t\| \, \|v_{x_j}\|}$ \Comment{Update image to maximize cosine similarity}
    \EndFor
    \State $S_i \gets \frac{v_t \cdot v_{x_m}}{\|v_t\| \, \|v_{x_m}\|}$ \Comment{Optimized similarity for epoch $i$}
    \State $\mathcal{S} \gets \mathcal{S} \cup \{S_i\}$, $\mathcal{V} \gets \mathcal{V} \cup \{v_{x_m}\}$
\EndFor
\State $S \gets \frac{1}{n} \sum_{S_i \in \mathcal{S}} S_i$
\State $\bar{v} \gets \frac{1}{n} \sum_{v \in \mathcal{V}} v$
\State $D \gets \sqrt{\frac{1}{n} \sum_{v \in \mathcal{V}} \|v - \bar{v}\|^2}$
\State \textbf{return} $S$, $D$
\end{algorithmic}
\end{algorithm}


\textbf{Generation of Textual Gibberish.} TUNI starts the detection process with generating a set of $\ell$ gibberish strings $\mathcal{G} = \{g_1,g_2,\ldots,g_{\ell}\}$, which are random combinations of digits and symbols with certain length. As these gibberish texts are randomly generated at the inference time, with overwhelming probability that they did not appear in the training set. Applying the proposed feature extraction algorithm on ${\cal G}$, we obtain $\ell$ feature vectors ${\cal F} = \{f_1,f_2,\ldots,f_{\ell}\}$ of the gibberish texts.

\textbf{Training Anomaly Detectors.} Motivated by the observations in Figure~\ref{fig:separation} that the feature vectors of the texts that are in and out of the training set of $M$ are well separated, we propose to train an anomaly detector using ${\cal F}$, such that texts out of $D_{\text{train}}$ are considered ``normal'', and the problem of ID inference on textual description $t$ is converted to anomaly detection on the feature vector of $t$. More specifically, $t$ is detected to be in $D_{\text{train}}$, if its feature vector is detected ``abnormal'' by the trained anomaly detector. Specifically in TUNI, we train several anomaly detection models on ${\cal F}$, such as Isolation Forest, LocalOutlierFactor~\citep{cheng2019outlier} and AutoEncoder~\citep{chandola2009anomaly}. These models constitute an anomaly detection voting system that will be used for ID inference on the test textual descriptions.



\textbf{Textual ID Inference through Voting.} For each textual description $t$ in the test set, TUNI first extracts its feature vector $f$ using Algorithm~\ref{alg:feature-extraction}, and then feeds $f$ to each of the obtained anomaly detectors to cast a vote on whether $t$ is an anomaly. When the total number of votes exceeds a predefined detetion threshold $N$, $t$ is determined as an anomaly, i.e., PII with textual description $t$ is used to train the CLIP model $M$; otherwise, $t$ is considered normal and no PII with $t$ is leaked through training of $M$. 


\begin{table*}[t]
\centering
\caption{\small Performance comparison with baseline methods across different CLIP models. $\Delta$ indicates the improvement of TUNI.}
\resizebox{\textwidth}{!}{
\begin{tabular}{@{}lcccccccccccc@{}}
\toprule
\textbf{Architecture} & \textbf{Number of photos per} & \textbf{Method} & \textbf{Precision} & $\Delta$ & \textbf{Recall} & $\Delta$ & \textbf{Accuracy} & $\Delta$ \\
 & \textbf{person in training set} & & & & & & & \\
\midrule
\multirow{6}{*}{ResNet-50} & \multirow{3}{*}{1} & WSA & 0.6653 ± 0.0032 & 0.1979 & 0.2925 ± 0.0045 & 0.6896 & 0.6675 ± 0.0037 & 0.2497 \\
 & & IDIA & 0.6922 ± 0.0023 & 0.1712 & 0.4032 ± 0.0027 & 0.5789 & 0.6836 ± 0.0034 & 0.2336 \\
 & & TUNI & \textbf{0.8634 ± 0.0031} & - & \textbf{0.9821 ± 0.0042} & - & \textbf{0.9172 ± 0.0028} & - \\
\cmidrule{2-9}
 & \multirow{3}{*}{75} & WSA & 0.6625 ± 0.0018 & 0.2017 & 0.2867 ± 0.0061 & 0.6968 & 0.6710 ± 0.0043 & 0.2322 \\
 & & IDIA & 0.6901 ± 0.0024 & 0.1741 & 0.3998 ± 0.0049 & 0.5837 & 0.6907 ± 0.0075 & 0.2125 \\
 & & TUNI & \textbf{0.8642 ± 0.0057} & - & \textbf{0.9835 ± 0.0019} & - & \textbf{0.9032 ± 0.0033} & - \\
\midrule
\multirow{6}{*}{ResNet-50x4} & \multirow{3}{*}{1} & WSA & 0.6712 ± 0.0029 & 0.1901 & 0.2912 ± 0.0048 & 0.6835 & 0.6808 ± 0.0031 & 0.2547 \\
 & & IDIA & 0.6625 ± 0.0036 & 0.1963 & 0.3980 ± 0.0031 & 0.5267 & 0.6957 ± 0.0029 & 0.2398 \\
 & & TUNI & \textbf{0.8613 ± 0.0033} & - & \textbf{0.9747 ± 0.0013} & - & \textbf{0.9355 ± 0.0038} & - \\
\cmidrule{2-9}
 & \multirow{3}{*}{75} & WSA & 0.6724 ± 0.0022 & 0.1988 & 0.2935 ± 0.0054 & 0.6981 & 0.6685 ± 0.0047 & 0.2777 \\
 & & IDIA & 0.7085 ± 0.0021 & 0.1627 & 0.3904 ± 0.0018 & 0.6012 & 0.7167 ± 0.0035 & 0.2295 \\
 & & TUNI & \textbf{0.8712 ± 0.0043} & - & \textbf{0.9916 ± 0.0037} & - & \textbf{0.9462 ± 0.0029} & - \\
\midrule
\multirow{6}{*}{ViT-B/32} & \multirow{3}{*}{1} & WSA & 0.6323 ± 0.0064 & 0.0268 & 0.2964 ± 0.0052 & 0.3421 & 0.6812 ± 0.0045 & 0.0025 \\
 & & IDIA & 0.6783 ± 0.0047 & 0.0308 & 0.3746 ± 0.0033 & 0.2639 & 0.6772 ± 0.0041 & 0.0065 \\
 & & TUNI & \textbf{0.7091 ± 0.0056} & - & \textbf{0.6385 ± 0.0062} & - & \textbf{0.6837 ± 0.0044} & - \\
\cmidrule{2-9}
 & \multirow{3}{*}{75} & WSA & 0.7045 ± 0.0075 & 0.0137 & 0.2806 ± 0.0048 & 0.3566 & 0.6895 ± 0.0052 & 0.0052 \\
 & & IDIA & 0.6890 ± 0.0051 & 0.0292 & 0.3811 ± 0.0063 & 0.2561 & 0.6927 ± 0.0045 & 0.0020 \\
 & & TUNI & \textbf{0.7182 ± 0.0068} & - & \textbf{0.6372 ± 0.0046} & - & \textbf{0.6947 ± 0.0078} & - \\
\bottomrule
\end{tabular}
}
\label{table:baseline}
\end{table*}

\textbf{Enhancement with Real Images.} At inference time, if real images of the test texts are available at the detector (e.g., photos of a person), they can be used to extract an additional feature measuring the average distance between the embeddings of real images and those of optimized images using the CLIP model, using which the feature vectors of the test texts can be clustered into two partitions with one in $D_{\text{train}}$ and another one out of $D_{\text{train}}$. This adds an additional vote for each test text to the above described anomaly detection voting system, potentially facilitating the detection accuracy.


Specifically, for each test text $t$, the detector is equipped with a set of $c$ real images $\{x_{\mathrm{real}}^1, x_{\mathrm{real}}^2, \ldots, x_{\mathrm{real}}^c\}$. Similar to the feature extraction process in Algorithm \ref{alg:feature-extraction}, over $k$ epochs with independent initializations, $k$ optimized images $\{x_{\mathrm{opt}}^1, x_{\mathrm{opt}}^2, \ldots, x_{\mathrm{opt}}^k\}$ for $t$ are obtained under the guidance of the CLIP model. Then, we apply a pretrained feature extraction model $F$ (e.g.,~DeepFace~\citep{taigman2014deepface} for face images) to the real and optimized images to obtain real embeddings $\{v_{\mathrm{real}}^1, v_{\mathrm{real}}^2, \ldots, v_{\mathrm{real}}^c\}$ and optimized embeddings $\{v_{\mathrm{opt}}^1, v_{\mathrm{opt}}^2, \ldots, v_{\mathrm{opt}}^k\}$. Finally, we compute average pair-wise $\ell_2$ distance between real and optimized embeddings, denoted by $R$, over $c \cdot k$ pairs, and use $R$ as an additional feature of the text $t$.


For a batch of $B$ test texts $(t_1,t_2,\ldots,t_B)$, we start with extracting their features $((S_1,D_1,R_1),(S_2,D_2,R_2),\ldots,(S_B,D_B,R_B))$. Feeding the first two features $S_i$ and $D_i$ into the trained anomaly detection system, each text $t_i$ obtains an anomaly score as the number of anomaly detectors who believe that it is abnormal. Additional, the $K$-means algorithm with $K=2$ is performed on the feature vectors $\{(S_i,D_i,R_i)\}_{i=1}^B$ to partition them into a ``normal'' cluster and an ``abnormal'' cluster, adding another vote on the anomaly score of each test instance. Then, the ID inference of each text is performed by comparing its total number of received votes and a detection threshold $N'$.

\section{Evaluations}
\label{sec:evalu}
We evaluate the performance of TUNI, for the task of ID inference from the name of a person, with the corresponding image being the face photo of the person.

\subsection{Setup}
Our experiments leverage datasets and target CLIP models from~\citep{hintersdorf2022}. 

\begin{figure}[h]
  \centering
  \includegraphics[width=0.48 \textwidth]{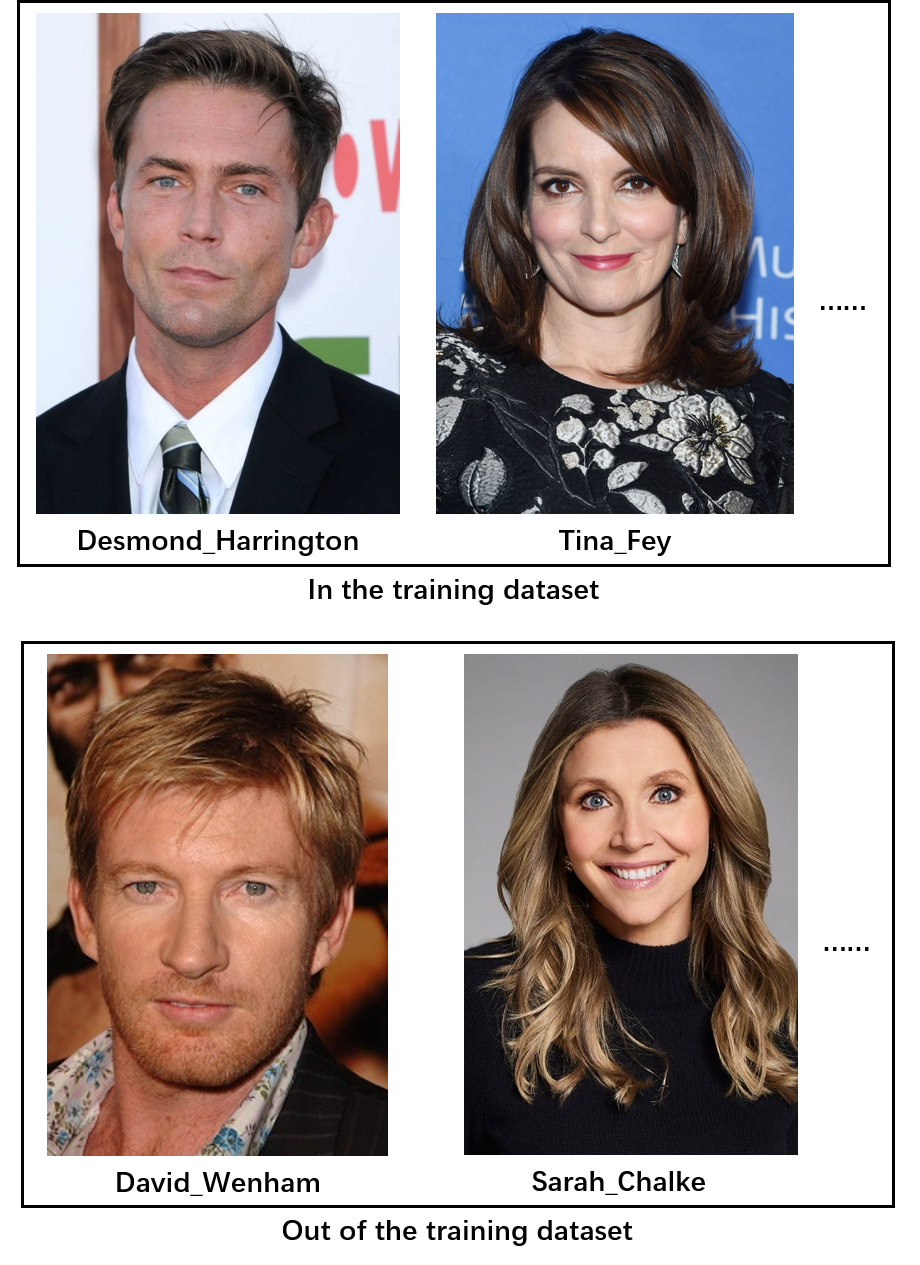}
  \caption{\small Samples from the dataset for training CLIP models.}
  \label{fig:pair}
\end{figure}

\textbf{Dataset Construction.} The datasets for training and ID inference are constructed from three datasets: LAION-5B~\citep{schuhmann2022laion}, Conceptual Captions 3M (CC3M)~\citep{changpinyo2021conceptual}, and FaceScrub~\citep{kemelmacher2016megaface}. Specifically, 200 celebrities—100 for training and 100 for validation, with their face photos accompanied by labels containing their names are selected from the FaceScrub dataset; then these data samples are augmented by additional photos of the selected celebrities found in LAION-5B, such that each person has multiple photos; finally these augmented  data points are mixed with the CC3M dataset to form the training set of the CLIP model. By doing this, we have the ground truth on which people are in the training set and which are not. In our experiments, we construct two datasets, one with a single photo for each person, and another with 75 photos for each person. Samples of this dataset are shown in Figure~\ref{fig:pair} and a more detailed description is given in appendix.

\textbf{Models.} Our analysis involves ID inference from six pre-trained target CLIP models, categorized into ResNet-50, ResNet-50x4, and ViT-B/32 architectures. The ResNet-50 and ResNet-50x4 models are based on the ResNet architecture~\citep{he2016deep,theckedath2020detecting}; and ViT-B/32 models employ the Vision Transformer architecture~\citep{chen2021vision}. DeepFace~\citep{serengil2020lightface} is used for facial feature extraction for enhancement with real images.

\textbf{Evaluation Metrics.} TUNI's effectiveness is assessed using Precision, Recall, and Accuracy metrics, measuring anomaly prediction accuracy, correct anomaly identification, and overall prediction correctness, respectively.

\textbf{Baselines.} Current ID inference detection methods for CLIP models typically require detector to query target model with corresponding real images. Most MIAs involve training shadow models and related methods like shadow encoders ~\citep{liu2021encodermi}, which can be particularly costly for large-scale multimodal models. We empirically compare the performance of TUNI with the following SOTA inference methods, which both avoid using shadow models, but still require submitting both text and image to the target CLIP model for inference.

\begin{itemize}
    \item {\bf Identity Inference Attack (IDIA)~\citep{hintersdorf2022}} detects with a list of 1000 names to choose from and 30 real photos for a tested person. In IDIA, the attacker (detector) selects candidate names as prompt templates, and predicts names for each image and prompt. Once the correct name is predicted, it's inferred that the target individual is in training dataset. We compare IDIA using 3 photos for each test sample with TUNI using only text.  
    \item {\bf Weakly Supervised Attack (WSA)~\citep{ko2023}} uses cosine similarity between image and text features to infer membership, and adds a weak supervision MIA framework based on non-member data generated after the release of the target model.
\end{itemize}

All experiments are performed using four NVIDIA GeForce RTX 3090 GPUs. Each experiment is repeated for 10 times, and the average values and the standard deviations are reported.

\subsection{Results} 
On training anomaly detectors, we randomly generated $\ell = 50$ textual gibberish (some of them are shown in Table~\ref{table:gibberish}).


The image optimization was performed for $n= 100$ epochs; and in each epoch, $m= 1000$ Gradient Descent (GD) iterations with a learning rate of $0.02$. Four anomaly detection models, i.e., LocalOutlierFactor~\citep{cheng2019outlier}, IsolationForest~\citep{liu2008isolation}, OneClassSVM~\citep{li2003improving,khan2014one}, and AutoEncoder~\citep{chen2018autoencoder} were trained, and $N=3$ was chosen as the detection threshold. 

As shown in Table~\ref{table:baseline}, TUNI, even with only text information, consistently outperforms WSA and IDIA in all metrics by a large margin, across all model architectures and datasets, demonstrating its superior performance. 



We also evaluate the effect of providing the TUNI detector with an real photo of the inferred person. In this case, the embedding distances between the real and optimized images of the test samples are used to perform a $2$-means clustering, adding another vote to the inference result. We accordingly raise the detection threshold $N'$ to $4$. As illustrated in Table~\ref{table:aphoto}, the given photo helps to improve the performance of TUNI across all tested CLIP models. While recalls in some ResNet models experience minor declines attributed to the raised threshold, all remain above 94\%. Conversely, the ViT-B models exhibit an almost 11\% increase in recall. A lower detection threshold aids recall enhancement but may concurrently lead to declines in other metrics. 

\begin{table*}[t]
\centering
\caption{\small Detection performance with a given photo during inference. $\Delta$ indicates performance improvement.}
\resizebox{0.9\textwidth}{!}{
\begin{tabular}{@{}lccccccccc@{}}
\toprule
\textbf{Architecture} & \multicolumn{1}{c}{\textbf{Number of photos per}} & \textbf{TUNI} & \textbf{Precision} & $\Delta$ & \textbf{Recall} & $\Delta$ & \textbf{Accuracy} & $\Delta$ \\
& \multicolumn{1}{c}{\textbf{person in training set}} & & & & & & & \\
\midrule
\multirow{4}{*}{ResNet-50} & \multirow{2}{*}{1} & Text only & 0.8634 ± 0.0031 & 0.1019 & {\bf 0.9821 ± 0.0042} & -0.0396 & 0.9172 ± 0.0028 & 0.0303 \\
 & & With 1 photo & \textbf{0.9653 ± 0.0032} & - & 0.9425 ± 0.0057 & - & \textbf{0.9475 ± 0.0041} & - \\
\cmidrule{2-9}
 & \multirow{2}{*}{75} & Text only & 0.8642 ± 0.0057 & 0.1183 & {\bf 0.9835 ± 0.0019} & -0.0188 & 0.9032 ± 0.0033 & 0.0538 \\
 & & With 1 photo & \textbf{0.9825 ± 0.0031} & - & 0.9467 ± 0.0024 & - & \textbf{0.9570 ± 0.0038} & - \\
\midrule
\multirow{4}{*}{ResNet-50x4} & \multirow{2}{*}{1} & Text only & 0.8613 ± 0.0033 & 0.1290 & {\bf 0.9747 ± 0.0013} & -0.0183 & 0.9355 ± 0.0038 & 0.0317 \\
 & & With 1 photo & \textbf{0.9923 ± 0.0011} & - & 0.9564 ± 0.0044 & - & \textbf{0.9672 ± 0.0028} & - \\
\cmidrule{2-9}
 & \multirow{2}{*}{75} & Text only & 0.8712 ± 0.0043 & 0.0912 & 0.9916 ± 0.0037 & 0.0019 & 0.9462 ± 0.0029 & 0.0323 \\
 & & With 1 photo & \textbf{0.9624 ± 0.0042} & - & \textbf{0.9935 ± 0.0029} & - & \textbf{0.9785 ± 0.0037} & - \\
\midrule
\multirow{4}{*}{ViT-B/32} & \multirow{2}{*}{1} & Text only & 0.7091 ± 0.0056 & 0.1432 & 0.6385 ± 0.0062 & 0.1084 & 0.6837 ± 0.0044 & 0.0975 \\
 & & With 1 photo & \textbf{0.8523 ± 0.0038} & - & \textbf{0.7469 ± 0.0078} & - & \textbf{0.7812 ± 0.0031} & - \\
\cmidrule{2-9}
 & \multirow{2}{*}{75} & Text only & 0.7182 ± 0.0068 & 0.1353 & 0.6372 ± 0.0046 & 0.1086 & 0.6947 ± 0.0078 & 0.1148 \\
 & & With 1 photo & \textbf{0.8535 ± 0.0042} & - & \textbf{0.7458 ± 0.0039} & - & \textbf{0.8095 ± 0.0063} & - \\
\bottomrule
\end{tabular}
}
\label{table:aphoto}
\vspace{-3mm}
\end{table*}

\subsection{Ablation Study} 
We further explore the impacts of different system parameters on the detection accuracy.

\begin{figure}[h]
\noindent
\begin{minipage}{0.24\textwidth}
  \includegraphics[width=\linewidth]{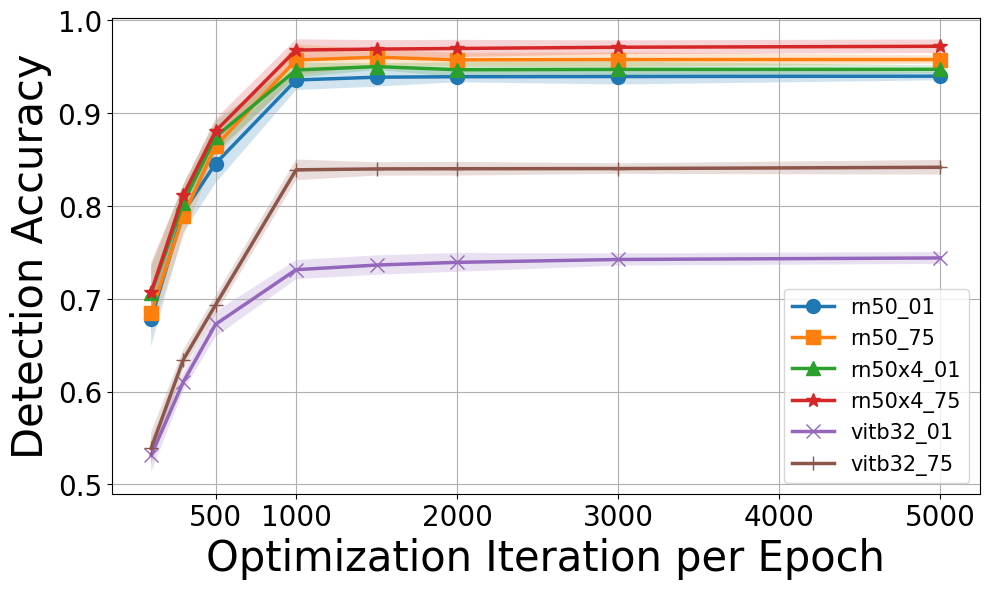}
  \caption{\small Detection accuracy for different numbers of optimization iterations per epoch.}
  \label{fig:iteration}
\end{minipage}%
\hspace*{4mm}
\begin{minipage}{0.24\textwidth}
  \includegraphics[width=\linewidth]{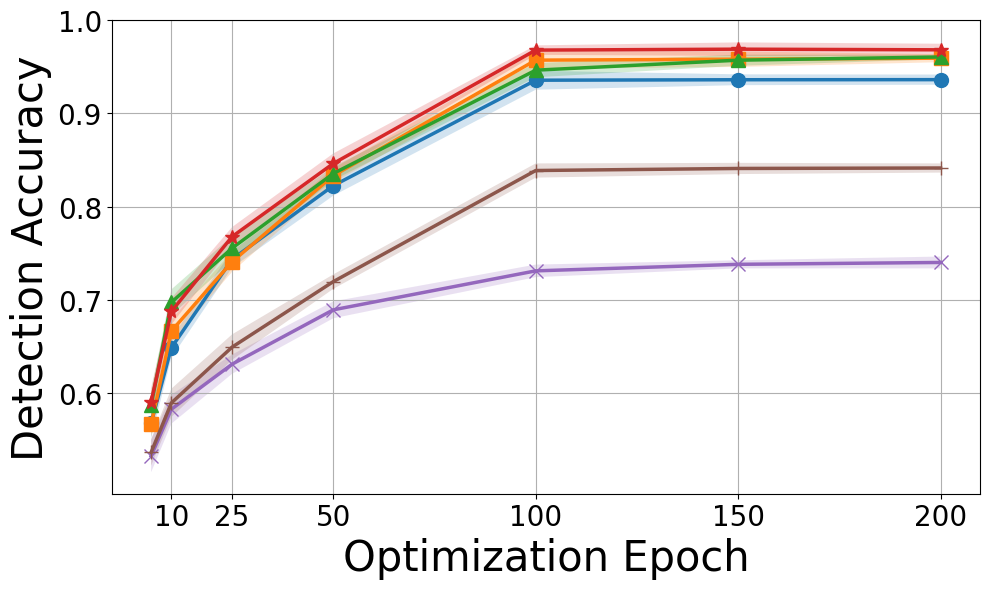}
  \caption{\small Detection accuracy for different numbers of epochs.}
  \label{fig:epoch}
\end{minipage}
\end{figure}

\begin{figure}[!h]
\noindent
\begin{minipage}{0.24\textwidth}
  \includegraphics[width=\linewidth]{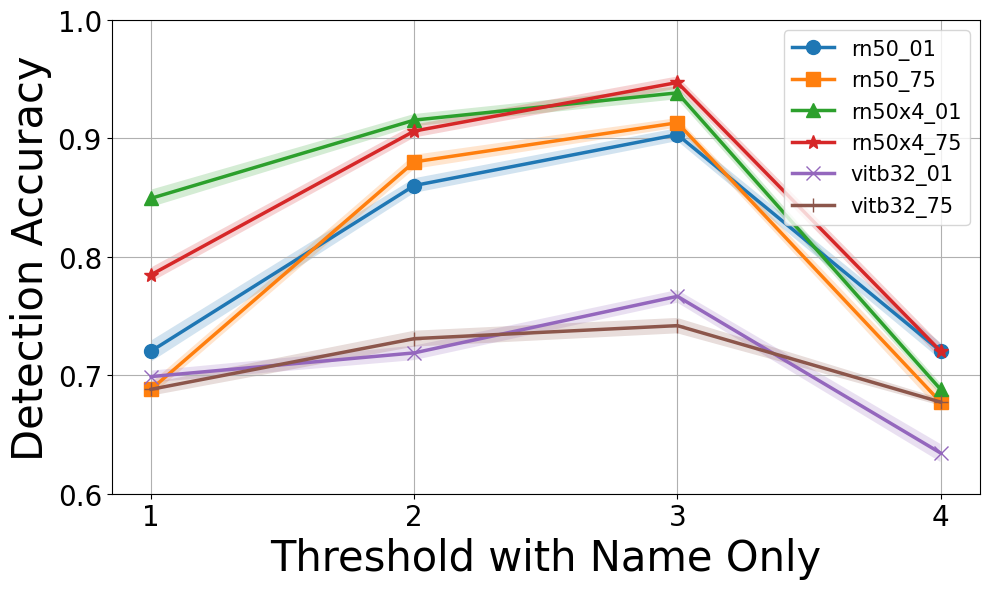}
  \caption{\small Detection accuracy with name only.}
\label{fig:textual}
\end{minipage}%
\hspace*{4mm}
\begin{minipage}{0.24\textwidth}
  \includegraphics[width=\linewidth]{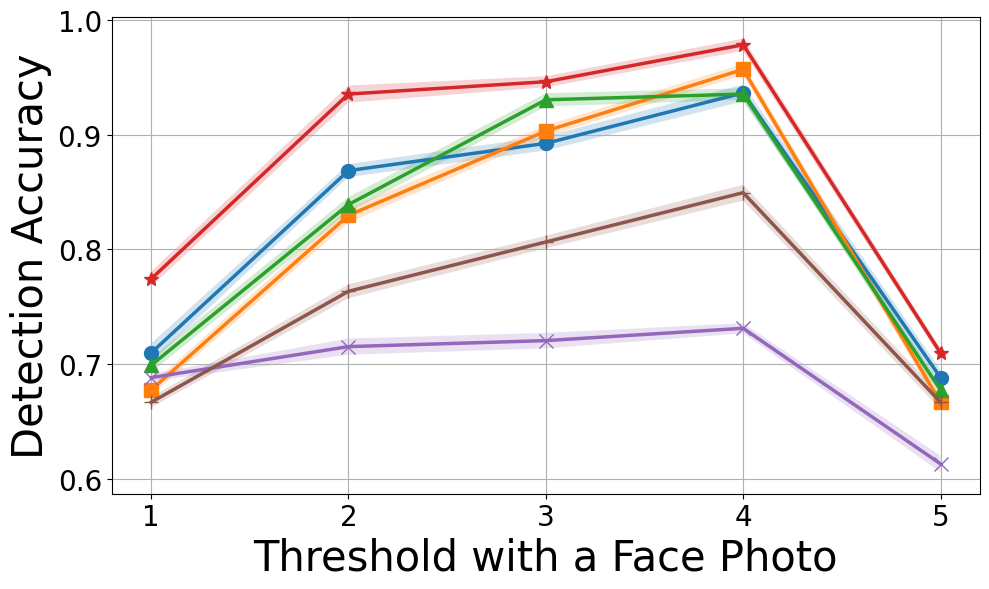}
  \caption{\small Detection accuracy with a face photo.}
   \label{fig:photo}
\end{minipage}
\end{figure}

\begin{figure}[!h]
\noindent
\begin{minipage}{0.24\textwidth}
  \includegraphics[width=\linewidth]{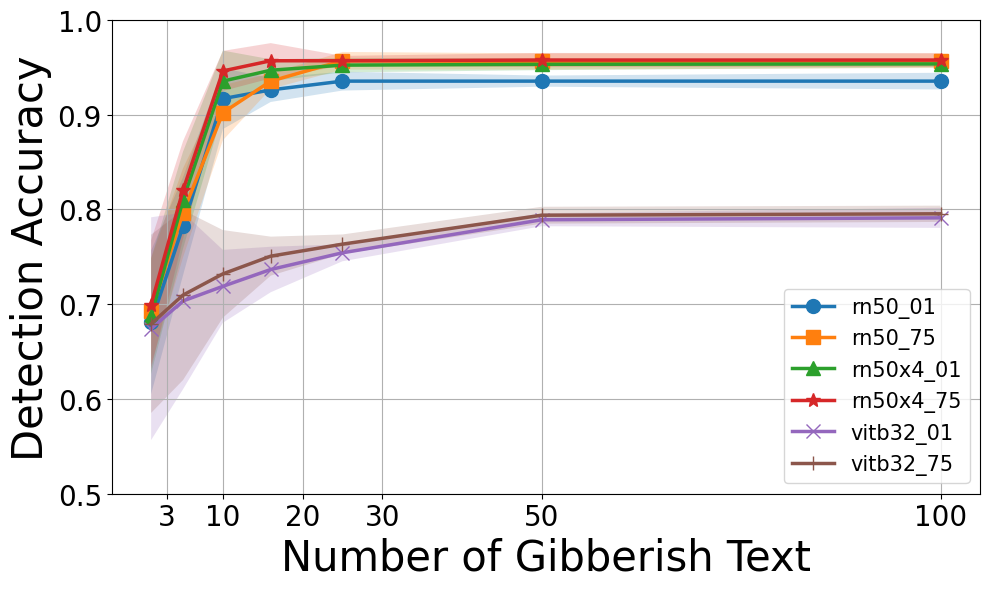}
  \caption{\small Detection accuracy for different numbers of gibberish.}
  \label{fig:gibberish}
\end{minipage}%
\hspace*{4mm}
\begin{minipage}{0.24\textwidth}
  \includegraphics[width=\linewidth]{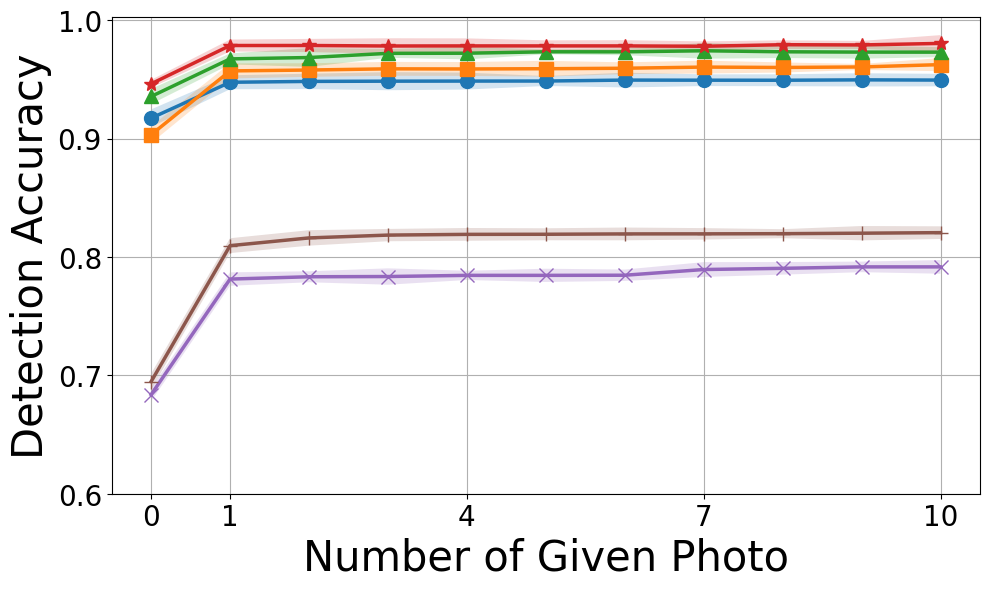}
  \caption{\small Detection accuracy for different number of real photos.}
  \label{fig:given photo}
\end{minipage}
\end{figure}

{\bf Optimization parameters.} Figure~\ref{fig:iteration} and \ref{fig:epoch} show that during feature extraction, optimizing for $n=100$ epochs, each with $m=1,000$ iterations, offers the optimal performance. Additional epochs and optimization iterations, while incurring additional computational cost, do not significantly improve the detection accuracy.

{\bf Detection threshold.} Figure~\ref{fig:textual} and \ref{fig:photo} show that the system attains higher accuracy, when it adopts a threshold of three votes for considering an input as an anomaly with text only, and four votes with an added detection model using an additional given photo. Setting a high threshold may result in failing to detect an anomaly, while setting a low one may lead to identifying a normal one as anomaly.

{\bf Number of textual gibberish.} As shown in Figure~\ref{fig:gibberish}, for different target models, the detection accuracies initially improve as the number of gibberish texts increases, and converge after using more than $50$ gibberish strings.



{\bf Number of real photos.} As shown in Figure~\ref{fig:given photo}, integrating real photos can enhance the detection accuracy; however, the improvements of using more than 1 photo are rather marginal.

\begin{table}[t]
\centering
\caption{\small Samples of randomly generated gibberish.}
\label{table:gibberish}
\begin{tabular}{|c|c|c|}
\hline
\b+7IKXb2Y & FR!pnI<5xS & euiT\_;\@yw/\\
\hline
je|\%5(s=G\textbackslash\_ & ?\^W<E\{Dvmz & hqf-~=j<q5 \\
\hline
\#lEZ0yrZ5ig & '2\_:6[jiOa & X*|<tFx|4/ \\
\hline
Fa<Z*Oike[ & \textbackslash 93W\~4>x5u & ?=\&QplxC-c \\
\hline
\end{tabular}
\end{table}

\section{Defense and Covert Gibberish Generation}
In real-world scenarios, target models being detected may deploy
defense mechanisms to 
recognize anomalous inputs like gibberish and provide misleading outputs, causing TUNI to misjudge inclusion of PII. 

To generate more covert gibberish data, we can create strings resembling normal text, with a few characters replaced by syllables from another language. For instance, the detector can craft query texts, by randomly combining English names with syllables from Arabic medical terminology. One way to do this is to start by prompting LLMs like GPT-3.5-turbo to create lists of common initial and final syllables in English words. These syllable lists are then extracted and refined to ensure diversity and eliminate duplicates. Next, the refined syllable combinations are randomly paired to create pseudo-English names, such as ``Karinix'', ``Zylogene'', ``Glycogenyx'', and ``Renotyl''. It's crucial to verify the novelty of these names by checking against a database of real names to avoid collision. Then by prompting the LLM to generate strings using the refined syllable combinations, covert gibberish strings resembling real names are produced (some examples are given in Table~\ref{table:covertgibberish}).

\begin{table}[t]
\centering
\caption{\small Covert gibberish that seem to be real names.}
\label{table:covertgibberish}
\begin{tabular}{|c|c|c|}
\hline
Karinix & Zylogene & Glycogenyx\\
\hline
Zylotrax & Vexilith & Dynatrix\\
\hline
Exodynix & Novylith & Glycosyne\\
\hline
Xenolynx & Rynexis & Delphylith\\
\hline
\end{tabular}
\end{table}

\section{Conclusion}
\label{sec:conclusion}
In this paper, we propose TUNI, the first method to conduct identity inference without exposing acutal images to target CLIP models. TUNI turns inference problem into anomaly detection, through randomly generating textual gibberish that are known to be out of training set, and exploting them to train anomaly detectors. 
Furthermore, the incorporation of real images is shown to enhance detection performance.
Through evaluations across various CLIP model architectures and datasets, we demonstrate the consistent superiority of TUNI over baselines.

\clearpage
\newpage

\section{Limitations} Due to constraints resources, we conducted experiments using the name of the individual as textual descriptions. This approach may not fully encapsulate the complexities and nuances of real-world PII leakage including addresses, phone numbers, and other sensitive information.

\section{Ethics and Social Impact}

The development of TUNI highlights crucial ethical considerations in identity inference using multimodal models like CLIP. By enabling identity inference with only textual data, TUNI reduces the risks associated with exposing PII through images. This approach not only helps protect individual privacy but also minimizes the potential for misuse in harmful applications. As such technologies evolve, it is essential for researchers to adhere to ethical guidelines and promote transparency, ensuring that advancements in AI prioritize user privacy and foster responsible usage in society.

\section{Potential Risks}
 TUNI aims to bolster privacy by aiding in identity inference and safeguarding personal identifiable information within AI systems. While mindful of the risk of misuse, TUNI should adhere to data regulations and be employed only with explicit consent from involved data subjects, promoting privacy and security in AI practices.

\bibliography{TUNI}

\begin{thebibliography}{68}
\providecommand{\natexlab}[1]{#1}

\bibitem[{Abadi et~al.(2016)Abadi, Chu, Goodfellow, McMahan, Mironov, Talwar,
  and Zhang}]{abadi2016deep}
Martin Abadi, Andy Chu, Ian Goodfellow, H~Brendan McMahan, Ilya Mironov, Kunal
  Talwar, and Li~Zhang. 2016.
\newblock Deep learning with differential privacy.
\newblock In \emph{Proceedings of the 2016 ACM SIGSAC conference on computer
  and communications security}, pages 308--318.

\bibitem[{Bonawitz et~al.(2017)Bonawitz, Ivanov, Kreuter, Marcedone, McMahan,
  Patel, Ramage, Segal, and Seth}]{bonawitz2017practical}
Keith Bonawitz, Vladimir Ivanov, Ben Kreuter, Antonio Marcedone, H~Brendan
  McMahan, Sarvar Patel, Daniel Ramage, Aaron Segal, and Karn Seth. 2017.
\newblock Practical secure aggregation for privacy-preserving machine learning.
\newblock In \emph{proceedings of the 2017 ACM SIGSAC Conference on Computer
  and Communications Security}, pages 1175--1191.

\bibitem[{Carlini et~al.(2021)Carlini, Tramer, Wallace, Jagielski,
  Herbert-Voss, Lee, Roberts, Brown, Song, Erlingsson
  et~al.}]{carlini2021extracting}
Nicholas Carlini, Florian Tramer, Eric Wallace, Matthew Jagielski, Ariel
  Herbert-Voss, Katherine Lee, Adam Roberts, Tom Brown, Dawn Song, Ulfar
  Erlingsson, et~al. 2021.
\newblock Extracting training data from large language models.
\newblock In \emph{30th USENIX Security Symposium (USENIX Security 21)}, pages
  2633--2650.

\bibitem[{Chandola et~al.(2009)Chandola, Banerjee, and
  Kumar}]{chandola2009anomaly}
Varun Chandola, Arindam Banerjee, and Vipin Kumar. 2009.
\newblock Anomaly detection: A survey.
\newblock \emph{ACM computing surveys (CSUR)}, 41(3):1--58.

\bibitem[{Changpinyo et~al.(2021)Changpinyo, Sharma, Ding, and
  Soricut}]{changpinyo2021conceptual}
Soravit Changpinyo, Piyush Sharma, Nan Ding, and Radu Soricut. 2021.
\newblock Conceptual 12m: Pushing web-scale image-text pre-training to
  recognize long-tail visual concepts.
\newblock In \emph{Proceedings of the IEEE/CVF conference on computer vision
  and pattern recognition}, pages 3558--3568.

\bibitem[{Chen et~al.(2021)Chen, Hsieh, and Gong}]{chen2021vision}
Xiangning Chen, Cho-Jui Hsieh, and Boqing Gong. 2021.
\newblock When vision transformers outperform resnets without pre-training or
  strong data augmentations.
\newblock \emph{arXiv preprint arXiv:2106.01548}.

\bibitem[{Chen et~al.(2018)Chen, Yeo, Lee, and Lau}]{chen2018autoencoder}
Zhaomin Chen, Chai~Kiat Yeo, Bu~Sung Lee, and Chiew~Tong Lau. 2018.
\newblock Autoencoder-based network anomaly detection.
\newblock In \emph{2018 Wireless telecommunications symposium (WTS)}, pages
  1--5. IEEE.

\bibitem[{Cheng et~al.(2019)Cheng, Zou, and Dong}]{cheng2019outlier}
Zhangyu Cheng, Chengming Zou, and Jianwei Dong. 2019.
\newblock Outlier detection using isolation forest and local outlier factor.
\newblock In \emph{Proceedings of the conference on research in adaptive and
  convergent systems}, pages 161--168.

\bibitem[{Cohen and Giryes(2024)}]{cohen2024membership}
Gilad Cohen and Raja Giryes. 2024.
\newblock Membership inference attack using self influence functions.
\newblock In \emph{Proceedings of the IEEE/CVF Winter Conference on
  Applications of Computer Vision}, pages 4892--4901.

\bibitem[{Dziedzic et~al.(2022)Dziedzic, Duan, Kaleem, Dhawan, Guan, Cattan,
  Boenisch, and Papernot}]{dziedzic2022dataset}
Adam Dziedzic, Haonan Duan, Muhammad~Ahmad Kaleem, Nikita Dhawan, Jonas Guan,
  Yannis Cattan, Franziska Boenisch, and Nicolas Papernot. 2022.
\newblock Dataset inference for self-supervised models.
\newblock \emph{Advances in Neural Information Processing Systems},
  35:12058--12070.

\bibitem[{Erlich et~al.(2018)Erlich, Shor, Pe’er, and
  Carmi}]{erlich2018identity}
Yaniv Erlich, Tal Shor, Itsik Pe’er, and Shai Carmi. 2018.
\newblock Identity inference of genomic data using long-range familial
  searches.
\newblock \emph{Science}, 362(6415):690--694.

\bibitem[{Golatkar et~al.(2022)Golatkar, Achille, Wang, Roth, Kearns, and
  Soatto}]{golatkar2022mixed}
Aditya Golatkar, Alessandro Achille, Yu-Xiang Wang, Aaron Roth, Michael Kearns,
  and Stefano Soatto. 2022.
\newblock Mixed differential privacy in computer vision.
\newblock In \emph{Proceedings of the IEEE/CVF Conference on Computer Vision
  and Pattern Recognition}, pages 8376--8386.

\bibitem[{Gu et~al.(2022)Gu, Meng, Lu, Hou, Minzhe, Liang, Yao, Huang, Zhang,
  Jiang et~al.}]{gu2022wukong}
Jiaxi Gu, Xiaojun Meng, Guansong Lu, Lu~Hou, Niu Minzhe, Xiaodan Liang, Lewei
  Yao, Runhui Huang, Wei Zhang, Xin Jiang, et~al. 2022.
\newblock Wukong: A 100 million large-scale chinese cross-modal pre-training
  benchmark.
\newblock \emph{Advances in Neural Information Processing Systems},
  35:26418--26431.

\bibitem[{He et~al.(2016)He, Zhang, Ren, and Sun}]{he2016deep}
Kaiming He, Xiangyu Zhang, Shaoqing Ren, and Jian Sun. 2016.
\newblock Deep residual learning for image recognition.
\newblock In \emph{Proceedings of the IEEE conference on computer vision and
  pattern recognition}, pages 770--778.

\bibitem[{He et~al.(2022)He, Li, Xu, Cornelius, and Zhang}]{he2022membership}
Xinlei He, Zheng Li, Weilin Xu, Cory Cornelius, and Yang Zhang. 2022.
\newblock Membership-doctor: Comprehensive assessment of membership inference
  against machine learning models.
\newblock \emph{arXiv preprint arXiv:2208.10445}.

\bibitem[{He and Zhang(2021)}]{10.1145/3460120.3484571}
Xinlei He and Yang Zhang. 2021.
\newblock \href {https://doi.org/10.1145/3460120.3484571} {Quantifying and
  mitigating privacy risks of contrastive learning}.
\newblock In \emph{Proceedings of the 2021 ACM SIGSAC Conference on Computer
  and Communications Security}, CCS '21, page 845–863, New York, NY, USA.
  Association for Computing Machinery.

\bibitem[{Helbling et~al.(2023)Helbling, Phute, Hull, and
  Chau}]{helbling2023llm}
Alec Helbling, Mansi Phute, Matthew Hull, and Duen~Horng Chau. 2023.
\newblock Llm self defense: By self examination, llms know they are being
  tricked.
\newblock \emph{arXiv preprint arXiv:2308.07308}.

\bibitem[{Hintersdorf et~al.(2022)Hintersdorf, Struppek, Brack
  et~al.}]{hintersdorf2022}
Daniel Hintersdorf, Lukas Struppek, Maximilian Brack, et~al. 2022.
\newblock Does clip know my face?
\newblock \emph{arXiv preprint arXiv:2209.07341}.

\bibitem[{Hisamoto et~al.(2020)Hisamoto, Post, and
  Duh}]{hisamoto2020membership}
Sorami Hisamoto, Matt Post, and Kevin Duh. 2020.
\newblock Membership inference attacks on sequence-to-sequence models: Is my
  data in your machine translation system?
\newblock \emph{Transactions of the Association for Computational Linguistics},
  8:49--63.

\bibitem[{Hu et~al.(2022{\natexlab{a}})Hu, Salcic, Sun, Dobbie, Yu, and
  Zhang}]{hu2022membership}
Hongsheng Hu, Zoran Salcic, Lichao Sun, Gillian Dobbie, Philip~S Yu, and Xuyun
  Zhang. 2022{\natexlab{a}}.
\newblock Membership inference attacks on machine learning: A survey.
\newblock \emph{ACM Computing Surveys (CSUR)}, 54(11s):1--37.

\bibitem[{Hu et~al.(2023)Hu, Yan, Yan, Li, Huang, Zhang, Dong, and
  Yang}]{hu2023defenses}
Li~Hu, Anli Yan, Hongyang Yan, Jin Li, Teng Huang, Yingying Zhang, Changyu
  Dong, and Chunsheng Yang. 2023.
\newblock Defenses to membership inference attacks: A survey.
\newblock \emph{ACM Computing Surveys}, 56(4):1--34.

\bibitem[{Hu et~al.(2022{\natexlab{b}})Hu, Wang, Sun, Wang, and Xue}]{hu2022m}
Pingyi Hu, Zihan Wang, Ruoxi Sun, Hu~Wang, and Minhui Xue. 2022{\natexlab{b}}.
\newblock M$^{4}$i: Multi-modal models membership inference.
\newblock \emph{Advances in Neural Information Processing Systems},
  35:1867--1882.

\bibitem[{Huang et~al.(2023)Huang, Liu, Nakada, Zhang, and
  Zhang}]{huang2023safeguarding}
Alyssa Huang, Peihan Liu, Ryumei Nakada, Linjun Zhang, and Wanrong Zhang. 2023.
\newblock Safeguarding data in multimodal ai: A differentially private approach
  to clip training.
\newblock \emph{arXiv preprint arXiv:2306.08173}.

\bibitem[{Inan et~al.(2021)Inan, Ramadan, Wutschitz, Jones, R{\"u}hle, Withers,
  and Sim}]{inan2021training}
Huseyin~A Inan, Osman Ramadan, Lukas Wutschitz, Daniel Jones, Victor R{\"u}hle,
  James Withers, and Robert Sim. 2021.
\newblock Training data leakage analysis in language models.
\newblock \emph{arXiv preprint arXiv:2101.05405}.

\bibitem[{Jagielski et~al.(2024)Jagielski, Nasr, Lee, Choquette-Choo, Carlini,
  and Tramer}]{jagielski2024students}
Matthew Jagielski, Milad Nasr, Katherine Lee, Christopher~A Choquette-Choo,
  Nicholas Carlini, and Florian Tramer. 2024.
\newblock Students parrot their teachers: Membership inference on model
  distillation.
\newblock \emph{Advances in Neural Information Processing Systems}, 36.

\bibitem[{Jia et~al.(2023)Jia, Liu, and Gong}]{jia202310}
Jinyuan Jia, Hongbin Liu, and Neil~Zhenqiang Gong. 2023.
\newblock 10 security and privacy problems in large foundation models.
\newblock In \emph{AI Embedded Assurance for Cyber Systems}, pages 139--159.
  Springer.

\bibitem[{Kalnis et~al.(2007)Kalnis, Ghinita, Mouratidis, and
  Papadias}]{kalnis2007preventing}
Panos Kalnis, Gabriel Ghinita, Kyriakos Mouratidis, and Dimitris Papadias.
  2007.
\newblock Preventing location-based identity inference in anonymous spatial
  queries.
\newblock \emph{IEEE transactions on knowledge and data engineering},
  19(12):1719--1733.

\bibitem[{Karaman and Bagdanov(2012)}]{karaman2012identity}
Svebor Karaman and Andrew~D Bagdanov. 2012.
\newblock Identity inference: generalizing person re-identification scenarios.
\newblock In \emph{Computer Vision--ECCV 2012. Workshops and Demonstrations:
  Florence, Italy, October 7-13, 2012, Proceedings, Part I 12}, pages 443--452.
  Springer.

\bibitem[{Kemelmacher-Shlizerman et~al.(2016)Kemelmacher-Shlizerman, Seitz,
  Miller, and Brossard}]{kemelmacher2016megaface}
Ira Kemelmacher-Shlizerman, Steven~M Seitz, Daniel Miller, and Evan Brossard.
  2016.
\newblock The megaface benchmark: 1 million faces for recognition at scale.
\newblock In \emph{Proceedings of the IEEE conference on computer vision and
  pattern recognition}, pages 4873--4882.

\bibitem[{Khan and Madden(2014)}]{khan2014one}
Shehroz~S Khan and Michael~G Madden. 2014.
\newblock One-class classification: taxonomy of study and review of techniques.
\newblock \emph{The Knowledge Engineering Review}, 29(3):345--374.

\bibitem[{Kim et~al.(2024)Kim, Yun, Lee et~al.}]{kim2024}
Seonghyeon Kim, Sooyeon Yun, Hwanil Lee, et~al. 2024.
\newblock Propile: Probing privacy leakage in large language models.
\newblock In \emph{Advances in Neural Information Processing Systems},
  volume~36.

\bibitem[{Ko et~al.(2023)Ko, Jin, Wang et~al.}]{ko2023}
Minseon Ko, Minseok Jin, Chen Wang, et~al. 2023.
\newblock Practical membership inference attacks against large-scale
  multi-modal models: A pilot study.
\newblock In \emph{Proceedings of the IEEE/CVF International Conference on
  Computer Vision}, pages 4871--4881.

\bibitem[{Leino and Fredrikson(2020)}]{leino2020stolen}
Klas Leino and Matt Fredrikson. 2020.
\newblock Stolen memories: Leveraging model memorization for calibrated
  $\{$White-Box$\}$ membership inference.
\newblock In \emph{29th USENIX security symposium (USENIX Security 20)}, pages
  1605--1622.

\bibitem[{Li et~al.(2003)Li, Huang, Tian, and Xu}]{li2003improving}
Kun-Lun Li, Hou-Kuan Huang, Sheng-Feng Tian, and Wei Xu. 2003.
\newblock Improving one-class svm for anomaly detection.
\newblock In \emph{Proceedings of the 2003 international conference on machine
  learning and cybernetics (IEEE Cat. No. 03EX693)}, volume~5, pages
  3077--3081. IEEE.

\bibitem[{Liang et~al.(2022)Liang, Liu, Liang, Li, Bai, and
  Cao}]{liang2022imitated}
Siyuan Liang, Aishan Liu, Jiawei Liang, Longkang Li, Yang Bai, and Xiaochun
  Cao. 2022.
\newblock Imitated detectors: Stealing knowledge of black-box object detectors.
\newblock In \emph{Proceedings of the 30th ACM International Conference on
  Multimedia}, pages 4839--4847.

\bibitem[{Liu et~al.(2008)Liu, Ting, and Zhou}]{liu2008isolation}
Fei~Tony Liu, Kai~Ming Ting, and Zhi-Hua Zhou. 2008.
\newblock Isolation forest.
\newblock In \emph{2008 eighth ieee international conference on data mining},
  pages 413--422. IEEE.

\bibitem[{Liu et~al.(2021)Liu, Jia, Qu, and Gong}]{liu2021encodermi}
Hongbin Liu, Jinyuan Jia, Wenjie Qu, and Neil~Zhenqiang Gong. 2021.
\newblock Encodermi: Membership inference against pre-trained encoders in
  contrastive learning.
\newblock In \emph{Proceedings of the 2021 ACM SIGSAC Conference on Computer
  and Communications Security}, pages 2081--2095.

\bibitem[{Liu et~al.(2022)Liu, Jia, Liu, and Gong}]{liu2022stolenencoder}
Yupei Liu, Jinyuan Jia, Hongbin Liu, and Neil~Zhenqiang Gong. 2022.
\newblock Stolenencoder: stealing pre-trained encoders in self-supervised
  learning.
\newblock In \emph{Proceedings of the 2022 ACM SIGSAC Conference on Computer
  and Communications Security}, pages 2115--2128.

\bibitem[{Lukas et~al.(2023)Lukas, Salem, Sim, Tople, Wutschitz, and
  Zanella-B{\'e}guelin}]{lukas2023analyzing}
Nils Lukas, Ahmed Salem, Robert Sim, Shruti Tople, Lukas Wutschitz, and
  Santiago Zanella-B{\'e}guelin. 2023.
\newblock Analyzing leakage of personally identifiable information in language
  models.
\newblock In \emph{2023 IEEE Symposium on Security and Privacy (SP)}, pages
  346--363. IEEE.

\bibitem[{Mattern et~al.(2023)Mattern, Mireshghallah, Jin, Sch{\"o}lkopf,
  Sachan, and Berg-Kirkpatrick}]{mattern2023membership}
Justus Mattern, Fatemehsadat Mireshghallah, Zhijing Jin, Bernhard
  Sch{\"o}lkopf, Mrinmaya Sachan, and Taylor Berg-Kirkpatrick. 2023.
\newblock Membership inference attacks against language models via
  neighbourhood comparison.
\newblock \emph{arXiv preprint arXiv:2305.18462}.

\bibitem[{Meeus et~al.(2023)Meeus, Jain, Rei, and de~Montjoye}]{meeus2023did}
Matthieu Meeus, Shubham Jain, Marek Rei, and Yves-Alexandre de~Montjoye. 2023.
\newblock Did the neurons read your book? document-level membership inference
  for large language models.
\newblock \emph{arXiv preprint arXiv:2310.15007}.

\bibitem[{Motahari et~al.(2009)Motahari, Ziavras, Schuler, and
  Jones}]{motahari2009identity}
Sara Motahari, Sotirios Ziavras, Richard~P Schuler, and Quentin Jones. 2009.
\newblock Identity inference as a privacy risk in computer-mediated
  communication.
\newblock In \emph{2009 42nd Hawaii International Conference on System
  Sciences}, pages 1--10. IEEE.

\bibitem[{Oh et~al.(2023)Oh, Park, Kim, Park, and Kwon}]{oh2023membership}
Myung~Gyo Oh, Leo~Hyun Park, Jaeuk Kim, Jaewoo Park, and Taekyoung Kwon. 2023.
\newblock Membership inference attacks with token-level deduplication on korean
  language models.
\newblock \emph{IEEE Access}, 11:10207--10217.

\bibitem[{Panda et~al.(2024)Panda, Choquette-Choo, Zhang, Yang, and
  Mittal}]{panda2024teach}
Ashwinee Panda, Christopher~A Choquette-Choo, Zhengming Zhang, Yaoqing Yang,
  and Prateek Mittal. 2024.
\newblock Teach llms to phish: Stealing private information from language
  models.
\newblock \emph{arXiv preprint arXiv:2403.00871}.

\bibitem[{Prince et~al.(2011)Prince, Li, Fu, Mohammed, and
  Elder}]{prince2011probabilistic}
Simon Prince, Peng Li, Yun Fu, Umar Mohammed, and James Elder. 2011.
\newblock Probabilistic models for inference about identity.
\newblock \emph{IEEE transactions on pattern analysis and machine
  intelligence}, 34(1):144--157.

\bibitem[{Radford et~al.(2021)Radford, Kim, Hallacy, Ramesh, Goh, Agarwal,
  Sastry, Askell, Mishkin, Clark et~al.}]{radford2021learning}
Alec Radford, Jong~Wook Kim, Chris Hallacy, Aditya Ramesh, Gabriel Goh,
  Sandhini Agarwal, Girish Sastry, Amanda Askell, Pamela Mishkin, Jack Clark,
  et~al. 2021.
\newblock Learning transferable visual models from natural language
  supervision.
\newblock In \emph{International conference on machine learning}, pages
  8748--8763. PMLR.

\bibitem[{Rahman(2023)}]{rahman2023survey}
Md~Abdur Rahman. 2023.
\newblock A survey on security and privacy of multimodal llms-connected
  healthcare perspective.
\newblock In \emph{2023 IEEE Globecom Workshops (GC Wkshps)}, pages 1807--1812.
  IEEE.

\bibitem[{Rahman et~al.(2024)Rahman, Alqahtani, Albooq, and
  Ainousah}]{rahman2024survey}
Md~Abdur Rahman, Lamyaa Alqahtani, Amna Albooq, and Alaa Ainousah. 2024.
\newblock A survey on security and privacy of large multimodal deep learning
  models: Teaching and learning perspective.
\newblock In \emph{2024 21st Learning and Technology Conference (L\&T)}, pages
  13--18. IEEE.

\bibitem[{Rahman et~al.(2020)Rahman, Fritz, Backes, and
  Zhang}]{rahman2020everything}
Tahleen Rahman, Mario Fritz, Michael Backes, and Yang Zhang. 2020.
\newblock Everything about you: A multimodal approach towards friendship
  inference in online social networks.
\newblock \emph{arXiv preprint arXiv:2003.00996}.

\bibitem[{Ramesh et~al.(2022)Ramesh, Dhariwal, Nichol, Chu, and
  Chen}]{ramesh2022hierarchical}
Aditya Ramesh, Prafulla Dhariwal, Alex Nichol, Casey Chu, and Mark Chen. 2022.
\newblock Hierarchical text-conditional image generation with clip latents.
\newblock \emph{arXiv preprint arXiv:2204.06125}, 1(2):3.

\bibitem[{Rigaki and Garcia(2023)}]{rigaki2023survey}
Maria Rigaki and Sebastian Garcia. 2023.
\newblock A survey of privacy attacks in machine learning.
\newblock \emph{ACM Computing Surveys}, 56(4):1--34.

\bibitem[{Sanderson and Lovell(2009)}]{sanderson2009multi}
Conrad Sanderson and Brian~C Lovell. 2009.
\newblock Multi-region probabilistic histograms for robust and scalable
  identity inference.
\newblock In \emph{Advances in biometrics: Third international conference, ICB
  2009, alghero, italy, june 2-5, 2009. Proceedings 3}, pages 199--208.
  Springer.

\bibitem[{Schuhmann et~al.(2022)Schuhmann, Beaumont, Vencu, Gordon, Wightman,
  Cherti, Coombes, Katta, Mullis, Wortsman et~al.}]{schuhmann2022laion}
Christoph Schuhmann, Romain Beaumont, Richard Vencu, Cade Gordon, Ross
  Wightman, Mehdi Cherti, Theo Coombes, Aarush Katta, Clayton Mullis, Mitchell
  Wortsman, et~al. 2022.
\newblock Laion-5b: An open large-scale dataset for training next generation
  image-text models.
\newblock \emph{Advances in Neural Information Processing Systems},
  35:25278--25294.

\bibitem[{Schuhmann et~al.(2021)Schuhmann, Vencu, Beaumont, Kaczmarczyk,
  Mullis, Katta, Coombes, Jitsev, and Komatsuzaki}]{schuhmann2021laion}
Christoph Schuhmann, Richard Vencu, Romain Beaumont, Robert Kaczmarczyk,
  Clayton Mullis, Aarush Katta, Theo Coombes, Jenia Jitsev, and Aran
  Komatsuzaki. 2021.
\newblock Laion-400m: Open dataset of clip-filtered 400 million image-text
  pairs.
\newblock \emph{arXiv preprint arXiv:2111.02114}.

\bibitem[{Schwartz and Solove(2011)}]{schwartz2011pii}
Paul~M Schwartz and Daniel~J Solove. 2011.
\newblock The pii problem: Privacy and a new concept of personally identifiable
  information.
\newblock \emph{NYUL rev.}, 86:1814.

\bibitem[{Serengil and Ozpinar(2020)}]{serengil2020lightface}
Sefik~Ilkin Serengil and Alper Ozpinar. 2020.
\newblock \href {https://doi.org/10.1109/ASYU50717.2020.9259802} {Lightface: A
  hybrid deep face recognition framework}.
\newblock In \emph{2020 Innovations in Intelligent Systems and Applications
  Conference (ASYU)}, pages 23--27. IEEE.

\bibitem[{Shamshad et~al.(2023)Shamshad, Naseer, and
  Nandakumar}]{shamshad2023clip2protect}
Fahad Shamshad, Muzammal Naseer, and Karthik Nandakumar. 2023.
\newblock Clip2protect: Protecting facial privacy using text-guided makeup via
  adversarial latent search.
\newblock In \emph{Proceedings of the IEEE/CVF Conference on Computer Vision
  and Pattern Recognition}, pages 20595--20605.

\bibitem[{Shokri et~al.(2017)Shokri, Stronati, Song, and
  Shmatikov}]{shokri2017membership}
Reza Shokri, Marco Stronati, Congzheng Song, and Vitaly Shmatikov. 2017.
\newblock Membership inference attacks against machine learning models.
\newblock In \emph{2017 IEEE symposium on security and privacy (SP)}, pages
  3--18. IEEE.

\bibitem[{Taigman et~al.(2014)Taigman, Yang, Ranzato, and
  Wolf}]{taigman2014deepface}
Yaniv Taigman, Ming Yang, Marc'Aurelio Ranzato, and Lior Wolf. 2014.
\newblock Deepface: Closing the gap to human-level performance in face
  verification.
\newblock In \emph{Proceedings of the IEEE conference on computer vision and
  pattern recognition}, pages 1701--1708.

\bibitem[{Theckedath and Sedamkar(2020)}]{theckedath2020detecting}
Dhananjay Theckedath and RR~Sedamkar. 2020.
\newblock Detecting affect states using vgg16, resnet50 and se-resnet50
  networks.
\newblock \emph{SN Computer Science}, 1(2):79.

\bibitem[{Truex et~al.(2019)Truex, Liu, Gursoy, Yu, and
  Wei}]{truex2019demystifying}
Stacey Truex, Ling Liu, Mehmet~Emre Gursoy, Lei Yu, and Wenqi Wei. 2019.
\newblock Demystifying membership inference attacks in machine learning as a
  service.
\newblock \emph{IEEE transactions on services computing}, 14(6):2073--2089.

\bibitem[{Wu et~al.(2022)Wu, Wen, Backes, Yu, and Zhang}]{wu2022model}
Yixin Wu, Rui Wen, Michael Backes, Ning Yu, and Yang Zhang. 2022.
\newblock Model stealing attacks against vision-language models.

\bibitem[{Xi et~al.(2024)Xi, Du, Li, Pang, Ji, Chen, Ma, and
  Wang}]{xi2024defending}
Zhaohan Xi, Tianyu Du, Changjiang Li, Ren Pang, Shouling Ji, Jinghui Chen,
  Fenglong Ma, and Ting Wang. 2024.
\newblock Defending pre-trained language models as few-shot learners against
  backdoor attacks.
\newblock \emph{Advances in Neural Information Processing Systems}, 36.

\bibitem[{Xue et~al.(2023)Xue, Yuan, He, Wu, Wu, Zhang, Liu, and Liu}]{9806361}
Mingfu Xue, Chengxiang Yuan, Can He, Yinghao Wu, Zhiyu Wu, Yushu Zhang, Zhe
  Liu, and Weiqiang Liu. 2023.
\newblock \href {https://doi.org/10.1109/TETC.2022.3184408} {Use the spear as a
  shield: An adversarial example based privacy-preserving technique against
  membership inference attacks}.
\newblock \emph{IEEE Transactions on Emerging Topics in Computing},
  11(1):153--169.

\bibitem[{Ye et~al.(2022)Ye, Maddi, Murakonda, Bindschaedler, and
  Shokri}]{ye2022enhanced}
Jiayuan Ye, Aadyaa Maddi, Sasi~Kumar Murakonda, Vincent Bindschaedler, and Reza
  Shokri. 2022.
\newblock Enhanced membership inference attacks against machine learning
  models.
\newblock In \emph{Proceedings of the 2022 ACM SIGSAC Conference on Computer
  and Communications Security}, pages 3093--3106.

\bibitem[{Yin et~al.(2021)Yin, Chen, Shou, and Chen}]{yin2021defending}
Yu~Yin, Ke~Chen, Lidan Shou, and Gang Chen. 2021.
\newblock Defending privacy against more knowledgeable membership inference
  attackers.
\newblock In \emph{Proceedings of the 27th ACM SIGKDD Conference on Knowledge
  Discovery \& Data Mining}, pages 2026--2036.

\bibitem[{Zhang et~al.(2023)Zhang, Wen, and Huang}]{zhang-etal-2023-ethicist}
Zhexin Zhang, Jiaxin Wen, and Minlie Huang. 2023.
\newblock \href {https://doi.org/10.18653/v1/2023.acl-long.709} {{ETHICIST}:
  Targeted training data extraction through loss smoothed soft prompting and
  calibrated confidence estimation}.
\newblock In \emph{Proceedings of the 61st Annual Meeting of the Association
  for Computational Linguistics (Volume 1: Long Papers)}, pages 12674--12687,
  Toronto, Canada. Association for Computational Linguistics.

\bibitem[{Zhou and Lam(2018)}]{zhou2018age}
Huiling Zhou and Kin-Man Lam. 2018.
\newblock Age-invariant face recognition based on identity inference from
  appearance age.
\newblock \emph{Pattern recognition}, 76:191--202.

\end{thebibliography}

\clearpage
\appendix

\section{Dataset Description}
\label{sec:dataset}

We utilized the datasets from previous work~\citep{hintersdorf2022}.

LAION-400M~\citep{schuhmann2021laion}, comprising 400 million image-text pairs, primarily employed for pre-training the CLIP model, offering a wide array of visual content and textual descriptions to facilitate the model's learning of relationships between images and text, including direct associations between specific individuals and images. In the experiment, this dataset is used to analyze the frequency of individuals appearing within it to identify individuals with lower frequencies of appearance, thereby avoiding the use of those individuals that appear very frequently to prevent skewing the experimental results. A threshold is set to only use individuals with fewer than 300 appearances for the experiments to ensure that the experimental results would not be dominated by individuals with very high occurrence frequencies, thus ensuring the accuracy and reliability of the experimental outcomes.

LAION-5B~\citep{schuhmann2022laion}, containing over 5.8 billion pairs and LAION-400M is its subset. In the experiment, LAION-5B is used to expand the CC3M dataset, enriching and increasing the sample size and diversity of the dataset. LAION-5B is used to find similar pairs to those in the FaceScrub dataset for each of the 530 celebrities. After confirming the presence of these celebrities' names in the captions of the found images, these image-text pairs were added to the CC3M dataset for training the target CLIP models.

Conceptual Captions 3M (CC3M)~\citep{changpinyo2021conceptual}, consisting of 2.8 million image-text pairs, anonymizes image captions by replacing named entities (e.g., celebrity names) with their hypernyms (e.g., "actor"). This dataset was also employed for pre-training the CLIP model. However, in this experiment, researchers analyzed the dataset using facial recognition technology to determine if specific celebrity images were present, and selectively added image-text pairs for model training adversarial attacks. As the named entities in CC3M dataset are anonymized in image captions, i.e., specific celebrity names replaced with their hypernyms like "actor," after confirming the presence or absence of specific celebrity images in the CC3M dataset, controlled additions of image-text pairs were made to the CC3M dataset.

FaceScrub~\citep{kemelmacher2016megaface}, containing images of 530 celebrities, was used to ascertain whether the identities one intends to infer are part of the training data. Celebrities were chosen due to the wide availability of their images in the public domain, minimizing privacy concerns associated with using their images.

To accurately calculate evaluation metrics, it was necessary to analyze which individuals were already part of the dataset and which were not. For the LAION-5B dataset, names of the 530 celebrities from the FaceScrub dataset were searched within all captions, and corresponding image-text pairs were saved, which were then added to the CC3M dataset. This was done to train the CLIP model and evaluate the effectiveness of IDIA under controlled conditions. In the experiments with the CC3M dataset, a total of 200 individuals were used, with 100 added to the dataset for model training and the remaining 100 held out for model validation. The selection of data in this process was balanced in terms of gender, with an equal distribution of male and female individuals to enhance the persuasiveness of the results. We construct two datasets for training the CLIP models of three architectures relatively, one with a single photo for each person, and another with 75 photos for each person. Samples of the datasets are shown in Figure~\ref{fig:pair}.






\end{document}